\documentclass[balance, nofoot, upint, pdf-a,colorlinks]{asmeconf}

\hypersetup{%
	pdfauthor={Kristen Edwards},
	pdftitle={ASME Conference Paper LaTeX Template},                  
	pdfkeywords={ASME conference paper, LaTeX template, BibTeX style},
	pdfsubject = {Describes the asmeconf LaTeX template},			  
	pdflicenseurl={https://ctan.org/pkg/asmeconf},
}

\usepackage{dblfloatfix}
\usepackage{enumitem}

\usepackage{graphicx}

\usepackage{amsmath}

\begin{document}



\ConfName{}
\ConfAcronym{}
\ConfDate{} 
\ConfCity{} 
\PaperNo{}


\title{ADVISE: AI-accelerated Design of Evidence Synthesis for Global Development}
 
%
%
%

\SetAuthors{%
	Kristen M.\ Edwards\affil{1}\JointFirstAuthor\CorrespondingAuthor{}, 
	Binyang Song\affil{1}\JointFirstAuthor,
        Jaron Porciello \affil{2},
        Mark Engelbert\affil{3}, 
        Carolyn Huang\affil{3},
	Faez Ahmed\affil{1}, 
	 }

\SetAffiliation{1}{Massachusetts Institute of Technology, Cambridge, MA }
\SetAffiliation{2}{University of Notre Dame, South Bend, IN}
\SetAffiliation{3}{International Initiative for Impact Evaluation, Inc.}



\maketitle



\keywords{AI in design, Natural Language Processing, Global Development, Evidence Synthesis}



\begin{abstract}
When designing evidence-based policies and programs, decision-makers must distill key information from a vast and rapidly growing literature base. Identifying relevant literature from raw search results is time and resource intensive, and is often done by manual screening. In this study, we develop an AI agent based on a bidirectional encoder representations from transformers (BERT) model and incorporate it into a human team designing an evidence synthesis product for global development. We explore the effectiveness of the human-AI hybrid team in accelerating the evidence synthesis process. To further improve team efficiency, we enhance the human-AI hybrid team through active learning (AL). Specifically, we explore different sampling strategies, including random sampling, least confidence (LC) sampling, and highest priority (HP) sampling, to study their influence on the collaborative screening process. Results show that incorporating the BERT-based AI agent into the human team can reduce the human screening effort by 68.5\% compared to the case of no AI assistance and by 16.8\% compared to the case of using a support vector machine (SVM)-based AI agent for identifying 80\% of all relevant documents. When we apply the HP sampling strategy for AL, the human screening effort can be reduced even more to 78.3\% for identifying 80\% of all relevant documents compared to no AI assistance. We apply the AL-enhanced human-AI hybrid teaming workflow in the design process of three evidence gap maps (EGMs) for USAID and find it to be highly effective. These findings demonstrate how AI can accelerate the development of evidence synthesis products and promote timely evidence-based decision making in global development in a human-AI hybrid teaming context. 

\end{abstract}

\section{Introduction}

In 2011 the U.S. Agency for International Development (USAID) released \textit{Evaluation Policy}, and in doing so made an ambitious commitment to rigorously evaluating evidence in order to make evidence-based policy~\cite{usaid2016strengthening}. Evidence-based policy refers to public policy that is based on, or informed by, evaluated and objective evidence. To emphasize the importance of evidence-based policy within USAID and the U.S. government, the Foundations for Evidence-based Policymaking Act of 2018 required all agencies under the Act to ``affirm the agency’s commitment to conducting rigorous, relevant, evaluations and to using evidence from evaluations to inform policy and practice''~\cite{congress2018evidencebasedpolicy}.
It is imperative in part because these policies dictate the expenditure of billions of dollars. For example, in 2017, USAID spent \$1.01 billion on foreign agricultural assistance alone~\cite{usaid2019agaid}.

However, evaluating all available evidence has been made burdensome by the current information explosion. In 2018 alone, global research output in science and engineering was 2.6 million articles, which grew at a rate of 4\% annually from 2008-2018~\cite{nsf}. A person's capacity to understand all available research is limited. Policy-makers have thus turned to evidence synthesis to understand the growing corpus of research available and make informed decisions. Evidence synthesis refers to the process of compiling information and knowledge from many sources and disciplines to inform decisions~\cite{donnelly2018egm, snilstveit_evidence_2016}. However, creating evidence synthesis products like evidence gap maps (EGMs) requires extensive time and effort from human experts. EGMs, as described in the Related Works section, visualize interventions and their associated outcomes~\cite{3ieEGM}, and have been shown to provide incredible value to decision-makers in fields ranging from agriculture to public health~\cite{donnelly2018egm}. For example, Figure~\ref{fig:exegm} represents a portion of an EGM available from 3ie \footnote{https://developmentevidence.3ieimpact.org/egm/food-systems-and-nutrition-evidence-gap-map}. We can see there is a research gap between the interventions of ``water access \& management'' and ``improved seeds'' and the outcomes regarding ``profit''. Policymakers can plan future investments and research accordingly.

Our goal is to accelerate the design of EGMs in the global development space and alleviate the burden of information filtering. The International Initiative for Impact Evaluation (3ie) is one of the global leaders in generating EGMs for decision-making. 3ie's current evidence synthesis process includes significant expert screening of documents and moderate use of machine intelligence, often taking nearly six months to complete~\cite{snilstveit_evidence_2016}. Natural language processing (NLP), a form of artificial intelligence (AI), has long been used for text comprehension. Recently, the rule-based NLP models have attracted some attention and been explored to promote evidence-based decision making in the medical, legal, and global development fields~\cite{biomedicus2019,bommarito2018lexnlp,porciello2020egm}. The work that has successfully done so may be improved upon by incorporating the latest transformer- and transfer learning-based NLP models.
\subsection{Contributions}
Title and abstract (TA) screening is one of the most time-consuming steps in the EGM design process, typically involving comprehending the titles and abstracts of tens or hundreds of thousands of papers for screening. Through collaborating with 3ie, we make the following contributions:
\begin{enumerate}[nolistsep]
    \item We develop a BERT-based AI agent to accelerate the TA screening portion of the EGM design process, and incorporate it into a human team to explore the efficiency gains made through human-AI teaming. With the best combination, our AI agent reduces human effort by 78.3\% when identifying 80\% of all eligible documents, as compared to no AI assistance. 
    \item We compare our BERT-based AI agent against the industry standard SVM-based model, and find that the BERT-based model outperforms SVM in both model performance (12\% average increase in accuracy for the three EGMs) and saved effort (17\% reduction in required effort in the simulated case, and a 46\% average reduction in effort for the three deployed EGMs).
    \item We identify the optimal training size (5,000 documents) for both model performance and saved effort.
    \item We compare active learning strategies and find that by using HP or LC we can decrease human effort by an additional 30\% (compared to BERT with no AL) for identifying 80\% of all included documents.
    \item We support the development of three EGMs: Agriculture, Nutrition, and Resilience. 
\end{enumerate}

\section{Related Work}

In the following sections we describe related work in the fields of evidence gap maps, natural language processing, and active learning, particularly in the context of human-AI teams. 

\subsection{Evidence Gap Maps}
EGMs are one form of evidence synthesis - the process of compiling information and knowledge from many sources and disciplines to inform decisions~\cite{donnelly2018egm, snilstveit_evidence_2016}. Evidence synthesis provides more reliable information about a topic than a single study by systematically collecting, categorizing, and analyzing a broad range of studies~\cite{briner_systematic_2012}. Evidence synthesis for decision making was largely popularized by the biomedical field, but it provides clear benefits for decision-makers in any field~\cite{sraboutsr2022,automationinsr2018, bommarito2018lexnlp}. Thus, evidence synthesis is an incredibly valuable tool for decision-makers in global development seeking to design policies and fund research~\cite{donnelly2018egm}. 

The International Initiative for Impact Evaluation (3ie) has pioneered the use of EGMs, which present a visual overview of completed and ongoing impact evaluations and systematic reviews in a specific sector~\cite{3iemapping}. 3ie creates these EGMs via the ``thematic [collection] of information about impact evaluation and systematic reviews that measure the effects of international development policies and programmes''~\cite{3ieEGM}. The final product is a matrix, organized by ``intervention'' categories on the vertical axis and ``outcome'' categories on the horizontal axis. Interventions are the action taken in the study, and outcomes are the result of the action. Each cell of the matrix contains studies that rigorously evaluate the impact of a specific intervention on a specific outcome.
\begin{figure}[htbp]
    \centering
    \includegraphics[width=\linewidth]{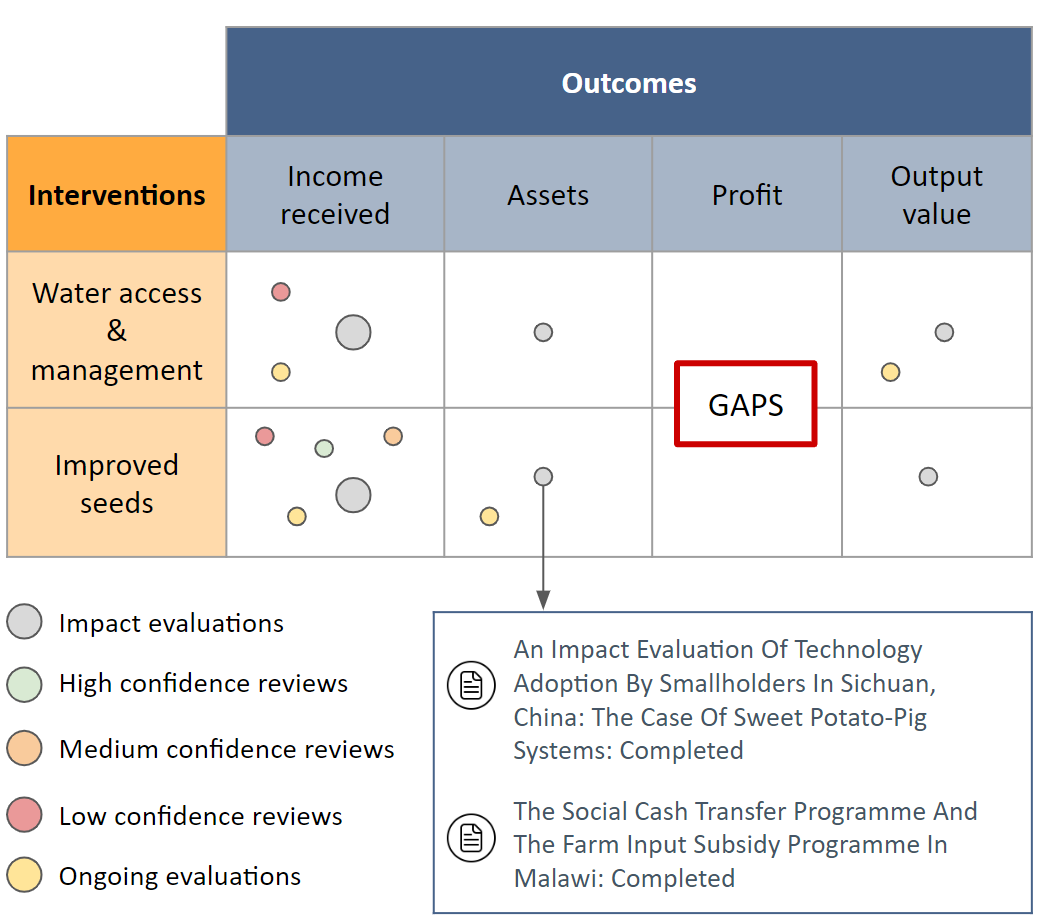}
    \caption{A representation of a portion of a 3ie EGM showing two interventions and four outcomes. Research gaps exist between the two interventions and the outcome ``Profit''. Dots of different colors represent different evidence types. Dot sizes indicate how many documents exist in each group.}
    \label{fig:exegm}
\end{figure}

\begin{figure*}[htbp]
    \centering
    \includegraphics[width=\textwidth]{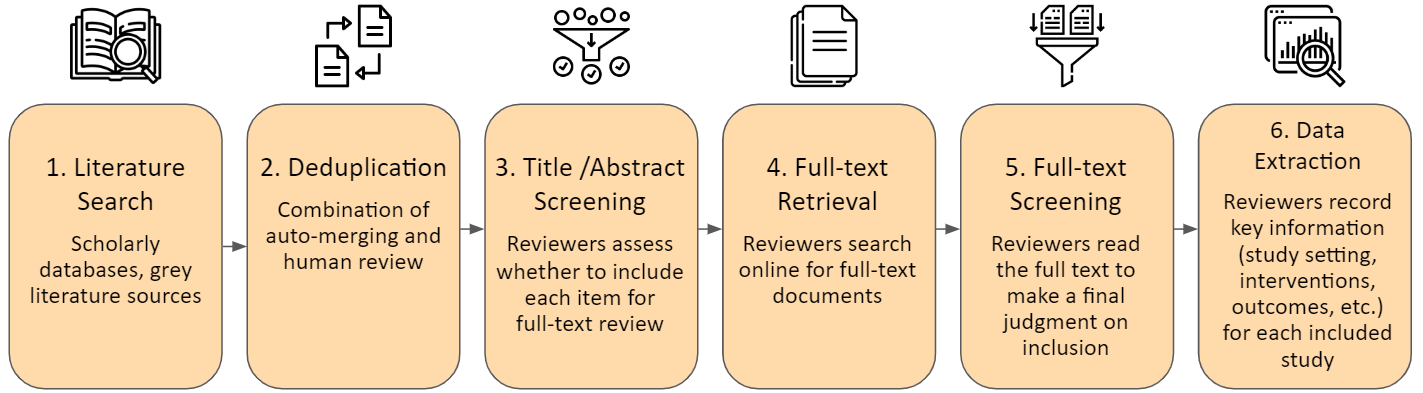}
    \caption{A high level view of the current EGM creation process.}
    \label{fig:egm}
\end{figure*}

3ie sets the global standard for EGMs, and the mapping method has been adapted by organizations including the Campbell Collaboration, the World Bank Independent Evaluation Group, and USAID~\cite{3iemapping}. Like other forms of evidence synthesis, EGMs begin with an expansive and systematic search of scholarly databases and ``grey literature'' sources (such as repositories of government documents or websites of think-tanks) to identify potentially relevant studies. EGM teams then screen these search results to identify studies that meet the EGM's criteria for interventions evaluated, outcomes measured, implementation setting, and study design. Once eligible studies are identified, the EGM team extracts information on interventions, outcomes, and other key characteristics of each study to determine its placement in the EGM matrix and to allow for analysis of trends in the literature. 

3ie uses a software called EPPI-Reviewer which aids in the creation of EGMs. While EPPI-Reviewer has some machine learning functions that can accelerate screening~\cite{omara-eves_using_2015}, most EGM tasks are still performed manually. Thus, each EGM requires significant human effort and expertise, with many EGMs requiring nearly six months to complete~\cite{snilstveit_evidence_2016}. Given that one of the main barriers to evidence use among policymakers is the lack of timely research outputs~\cite{oliver_systematic_2014}, there is a critical need to reduce the time and effort needed to complete the EGM design and development process.

The high level steps of designing an EGM are shown in Figure~\ref{fig:egm}. Our work focuses on step three, in which reviewers screen documents for inclusion in an EGM based on their title and abstract. Selected documents will move on to full-text review. We create three transformer-based NLP models that automatically classify documents for inclusion at this step. 

\subsection{Natural Language Processing in Evidence Synthesis} 

NLP is a field of machine learning in which computational machines are trained to understand text and spoken language.  Historically in NLP words are represented as vectors where similar words are located near each other in continuous space~\cite{word2mikolov2013distributed, word2mikolov2013efficient, pennington2014glove}. In the medical field, the development of an NLP-based model for automating evidence synthesis, called BioMedICUS, improved the scalability and performance of text analysis and processing of biomedical and clinical reports~\cite{biomedicus2019}. The success of NLP in the medical field has led to its use in other fields, with models like LexNLP, which automatically extracts information from legal text~\cite{bommarito2018lexnlp}.

There are several industry-standard NLP tools used to aid human experts when designing evidence synthesis products like EGMs. The most common tools include EPPI-Reviewer, Rayyan, and RobotReviewer~\cite{sraboutsr2022, automationinsr2018}, and all of these utilize support vector machines (SVM) as their primary ML model~\cite{rayyan2016, robotreviewer2017, THOMAS2021140}.

Furthermore, rule-based NLP models have been explored to promote evidence-based decision making in multiple fields~\cite{porciello2020egm}. However, the rule-based models are often case-specific. It requires significant effort to adapt a rule-based model from one EGM to another. Moreover, it is challenging to capture all the subjective criteria used by humans and embed them into the defined rules comprehensively. 


Modern NLP models have been largely shaped by the introduction of the transformer in 2017, which allowed text inputs to be fed in parallel and achieved state-of-the-art results over SVM and other models in many NLP tasks~\cite{vaswani2017attention}. Bidirectional Encoder Representations from Transformers (BERT) is among the most well-known transformer-based models and has been extensively explored in NLP tasks such as language translation and question answering~\cite{devlin2019bert, liu2019roberta}. Other such models include GPT and models based off of it~\cite{radford2019language}.

Our work explores BERT-based NLP models as a tool for human-AI teams designing EGMs for global development, which involves more unstructured studies and broader domains than other fields. Recent research that is perhaps most similar to our work is srBERT~\cite{srBERT2021}, which explores fine-tuning a BERT model with topic-specific articles in order to accelerate the screening process for a systematic review about ``moxibustion for improving cognitive impairment''~\cite{srBERT2021}. Our work, on the other hand, works with much larger and broader datasets in order to create EGMs. Additionally, we exhibit the effectiveness of our NLP tool in a real human-AI team and ultimately create three deployed EGMs in Agriculture, Nutrition, and Resilience. We utilize the experience of creating deployed EGMs to explore the nuances of human-AI teaming in this design process.

\subsection{Active Learning and Human-AI Teams} In many AI tasks, obtaining labeled training data is expensive and time-consuming~\cite{ALBERT}. We are motivated to explore avenues to decrease the size of training data needed by using active learning (AL). AL is the concept that an ML algorithm can perform better with less training data if it is allowed to choose the data from which it learns. AL has been applied to deep learning problems such as image classification~\cite{Aggarwal_2021, gal2017}, speech recognition~\cite{tur2005}, data exfiltration detection~\cite{dataexfiltration}, and many NLP tasks~\cite{olsson2009}. There are three main problem setups, or scenarios, in which a learner may be able to ask queries: membership query synthesis~\cite{angluin1998}, stream-based selective sampling~\cite{cohn1989}, and pool-based sampling\cite{lewis1994}. The most commonly used pool-based sampling strategies evaluate and rank the entire unlabeled pool in terms of informativeness and then select the best queries~\cite{alsurvey}. There are also different \textit{query strategies} for choosing which unlabeled instances to query. The most commonly used query strategy is uncertainty sampling~\cite{lewis1994}. In this strategy, the learner queries the instances for which the learner is least certain how to label~\cite{alsurvey}. Within the uncertainty sampling category, there are three primary measures that evaluate how uncertain the learner is about each instance: least confidence~\cite{leastconfidence}, margin sampling~\cite{Scheffer2001ActiveHM}, and entropy~\cite{shannon1948}.

During the training, AL tries to optimize the information flow from humans to AI to improve AI performance with less training data. In this study, AI is working collaboratively with humans instead of alone. Accordingly, both the information flow from humans to AI and that from AI to humans are important to the performance of human-AI hybrid teams. Therefore, AL in such a context should consider the bi-directional information flows.

In fact, common barriers stopping human screeners from incorporating AI into their EGM design process include a mismatch in existing workflows, and a steep learning curve~\cite{automationinsr2018}. Research shows that human-AI teams using active learning for a real life task can lose the human agent's trust if the AI agent makes irrelevant suggestions or predictions during the training process~\cite{dataexfiltration}.  Therefore, we explore and discuss the effective integration of AI tools into the existing EGM design process. We also use AL-based approaches to maximize the accuracy of the AI classifier, while minimizing the workload put on the human screeners. 
\begin{figure*}[h]
    \centering
    \includegraphics[width=\textwidth]{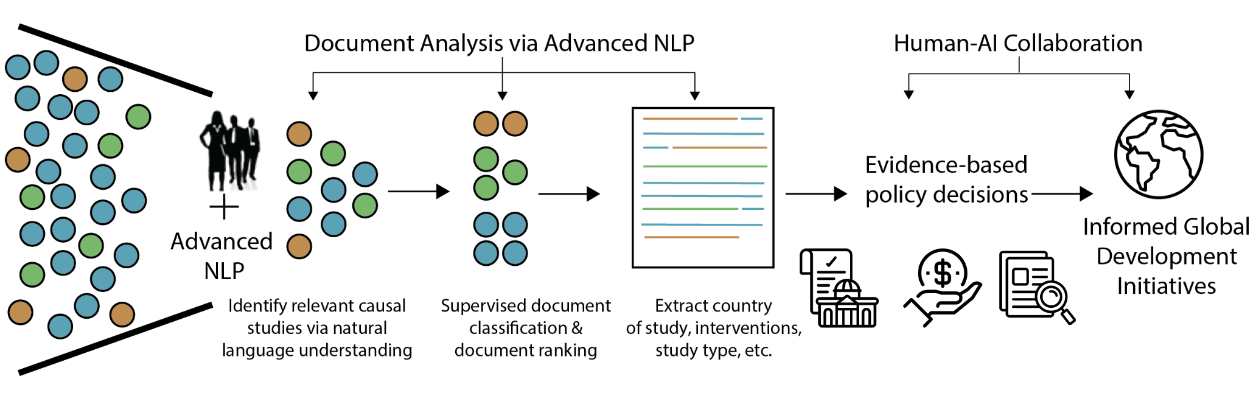}
    \caption{Proposed utilization of NLP tools in a human-AI team to screen, understand, and classify documents (represented by circles) in order to inform evidence-based policy decisions. Our goal is to accelerate the design process for EGM products in the global development field.}
    \label{fig:human-AI}
\end{figure*}

\section{Methodology}
In this work, we utilize a BERT-based NLP model to accelerate the design process of evidence gap maps. We utilize the NLP tool in the workflow of a real human-AI team with members of 3ie. We supported the development of three deployed EGMs in the topics of Agriculture, Nutrition, and Resilience.

We analyze how different ML methods and data training sizes affect the human-AI team performance, focusing on the trade-offs between model accuracy and human effort. We compare our fine-tuned BERT model against the industry standard SVM-based NLP tools. Further, we explore the effect of active learning with various query strategies on model accuracy and human effort. 

Our work is comprised of two case studies:
\begin{enumerate}
    \item \textbf{Deployed EGM design}: Actively designing and creating three EGMs regarding Agriculture, Nutrition, and Resilience. 
    \item \textbf{Simulated EGM design}: Using a pre-existing, fully labeled dataset to retrospectively study the most effective classification algorithms (SVM vs. BERT) and active learning strategies for EGM creation. 
\end{enumerate}

These two case studies present different challenges and priorities. In the simulated EGM design, we have the benefit of a fully labeled dataset, which we can practice multiple techniques on. In the deployed EGM design, we are in the real-world situation of creating an EGM from scratch using a human-AI team. Therefore, we only have labels for the documents that we specifically choose to screen. 

Further, in the deployed EGM design, we are motivated to design the most comprehensive and informative EGM while efficiently utilizing human resources. Consequently, we want to minimize time that human experts spend screening irrelevant documents, and screen only the relevant documents. This contrasts the strategy in classical active learning to query or screen the documents we are most uncertain of.

\subsection{Dataset Description}
For the simulated EGM design, our data is provided by 3ie and is derived from manually labeled documents from 3ie's Development Evidence Portal (DEP) \footnote{https://developmentevidence.3ieimpact.org/}~\cite{3ieDEP}, an expansive repository of impact evaluations and systematic reviews in global development across a wide range of sectors.  We utilize a dataset of 68,539 documents screened for inclusion in 3ie's DEP to develop and evaluate our classification model. Table~\ref{table:dataattributes} shows the key attributes of the dataset, such as title, abstract, and inclusion decisions.

\begin{table}[ht]
\centering
\begin{tabular}{p{2.6cm} p{4.9cm}}
\hline
\textbf{Attribute} & \textbf{Description} \\
\hline
Title & Title of the paper. \\
Abstract & Abstract of the paper. \\
Keywords & Keywords of the paper. \\
Year & Publication year. \\
Publication type & Journal, conference proceeding, report, etc. \\
Source & The source of the paper, e.g., journals or conferences. \\
Inclusion decision & Whether the paper is included as a relevant study. If not, what is the exclusion criterion.\\
\hline
\end{tabular}
\caption{The key attributes in the Development Evidence Portal dataset.}
\label{table:dataattributes}
\end{table}
In this study, the title of each paper is integrated into the abstract as a sentence at the beginning. The BERT classification model takes the integrated texts as the input. The label of ``included'' or ``excluded'' is derived from the inclusion decision. To train the binary classification model, the ``0'' class corresponds to the ``excluded'' papers, and the ``1'' class comprises the ``included'' papers. This dataset is highly imbalanced, containing 5,281 included papers and 63,258 excluded papers. The criteria for excluding the papers are also extracted for training the criterion-specific classification models.

For our deployed EGM design, we are actively designing three EGMs. As indicated in Figure~\ref{fig:egm}, and per 3ie's EGM workflow, we gather our initial dataset via a literature search through scholarly databases and grey literature sources. For the three EGMs, their initial dataset sizes are as follows: Agriculture 221k, Nutrition 117k, and Resilience 60k. 

\subsection{Data Pre-processing}

The raw documents are pre-processed to remove noise. Two types of noise are removed in this step. The first is non-English texts. A portion of the papers provide titles and abstracts in multiple languages. Since our models only take texts in English as input, the sentences in languages other than English are noise to the models and should be removed. The second type comprises English text content that is irrelevant to the scope of the document, such as a copyright statement. The pre-processing consists of five steps. (1) Each document is parsed into sentences. (2) A language detection model is used to identify sentences written in non-English languages. (3) We manually label the sentences from 500 documents with the ``relevant'' and ``irrelevant'' labels. (4) A BERT classification model is trained on the labeled data to predict the labels of the other sentences. The accuracy of the model is higher than 0.99. (5) Once the irrelevant sentences are removed, the remaining relevant sentences are integrated back into the original documents.

\subsection{Priority Score}

In this study, the AI agent is operationalized by a BERT binary classification model, which employs a 12-layer pre-trained uncased BERT embedding module with a hidden size of 768. The BERT embedding module is followed by a dropout layer with a drop rate = 0.1 and a linear layer that outputs a 2-dimensional (2D) vector as the final classification prediction. As described above, the AI agent needs to sample or prioritize the unlabeled papers according to the probabilities of being relevant, as predicted by the classification model. This probability is named the ``\emph{priority score}'' (PS) in definition 1.

\textbf{Definition 1.} \emph{Priority score} is the probability that a paper is a relevant paper predicted by the AI agent, which is calculated by $PS(p)=softmax(Pred(p))[1]$, where $Pred(p)$ is the prediction output from the classification model for a paper $p$, which is a 2D vector. The ``1'' in the equation indicates that PS(p) is the probability of the paper being classified to the ``1'' class. Following this definition, higher screening priority scores are assigned to the papers with higher predicted probabilities of being relevant.


\subsection{Sampling Strategies}

According to the predicted PSs, we apply three different query strategies to sample papers from the unscreened subset, which will be labeled and added to the training set in the next iteration.


\begin{enumerate}
    \item \textbf{Least confidence}: The \emph{least confidence (LC)} query strategy is one of the commonly used strategies for active learning, which samples papers that the model is least certain how to classify~\cite{alsurvey}, as shown by 
    \begin{equation}
        x_{LC}^{*}=\operatorname*{arg\,max}_{x\in X} U(x)
    \end{equation}
    The classification uncertainty $U(p)$ of the paper \textit{p} is derived from the classification model output $Pred(p)$ through 
    \begin{equation}
        U(p)=1-max(softmax((Pred(p)))
    \end{equation}
    \item \textbf{Highest priority}: The \emph{highest priority (HP)} query strategy samples papers with the highest PSs, given by 
    \begin{equation}
        x_{HP}^{*}=\operatorname*{arg\,max}_{x\in X}PS(x)
    \end{equation}
    This query strategy is adapted from the uncertainty sampling strategies~\cite{lewis1994}. For evidence synthesis, all relevant papers need to be verified by a human agent, so the papers most likely to be relevant are first sampled.
    \item \textbf{Random}: The \emph{random} query strategy randomly samples papers from the unlabeled list without using any informativeness measure. In this case, no AL is applied.
 \end{enumerate}







\subsection{Human-AI Hybrid Team Workflow} \label{humanaiteam}

We assume the human-AI hybrid team is tasked with screening a set of papers to identify papers satisfying a given scope. The human agents start the TA screening process by specifying the screening criteria. Then, the AI agent randomly samples a subset of papers, which are screened by the human agents as the initial training set. On this basis, our model is trained to learn the screening criteria from the training. With the learned knowledge, the AI agent predicts the PSs of the unscreened papers. According to the predicted PSs, the AI agent needs to check whether the screen-train-predict-sample loop should stop. If not, it employs a certain strategy to sample a set of papers to be screened for the next iteration. 

\begin{figure}[!htb]
\centering
\includegraphics[width=\columnwidth]{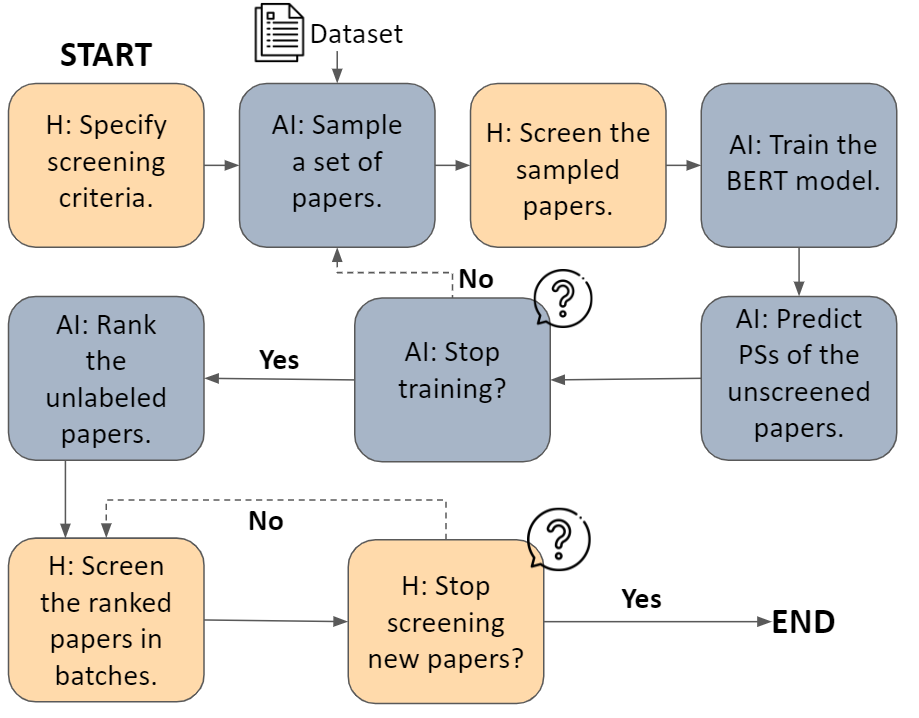}
\caption{The workflow of the human-AI hybrid team. H~represents human agents.}
\label{fig:work}
\end{figure}

Once the AI agent decides to stop training the model after a few iterations, it prioritizes all the unscreened papers according to their predicted PSs. Then, the human agents screen the prioritized papers in batches and decide at which batch the screening process should be ended. Since the dataset is imbalanced, a random over-sampling method is applied to the training set to make the numbers of samples from both classes equal. In this study, the AI agent samples a batch of 1,000 papers each time. The batch number is selected because it balances the gain from and the cost of updating the model.

\subsection{Evaluation Metrics}\label{evaluation metrics}

In the ML domain, accuracy and F1 score are commonly used to evaluate the performance of classification models. However, these metrics alone are not informative enough to assess the performance and efficiency of the human-AI hybrid teams. In this study, we evaluate the performance of the human-AI hybrid teams in terms of \emph{human effort} in definition 2 needed for achieving an \emph{inclusion rate} in definition 3. The computational cost reflects team efficiency from another perspective, which is not discussed in this study.

\textbf{Definition 2.} \emph{Human effort} is defined as the ratio ($HE$) between the number of papers that need to be screened manually ($n_{screened}$) for identifying a given amount of relevant papers and the total number of papers ($n=68,539$ in the simulated EGM design case study) in the dataset, which is calculated by: $HE= {n_{screened}}/{n}$.

\textbf{Definition 3.} \emph{Inclusion rate} is the ratio ($IR$) between the number of included papers being identified ($n_{identified}$) and the total number of included paper ($n_{included}=5,281$ in the simulated EGM design case study) in the dataset, calculated by: $IR={n_{identified}}/{n_{included}}$. With limited resources, a higher inclusion rate is preferred.

Given a set of scientific papers and the screening criteria, an efficient human-AI hybrid team should minimize the human effort and computational cost for achieving a satisfying inclusion rate or maximize the inclusion rate with available human effort and computational resources. Additionally, the F1 score of the corresponding classification model in each case is also reported for assessing the performance of the AI agent.
\section{Experimental Setup}
In this section, we discuss the experimental setups for the two case studies: deployed EGM design and simulated EGM design. The aspects specific to each case study are described in~\ref{deployedexperiment} and~\ref{simulatedexperiment}, while their shared components like baseline techniques and implementation details are described in sections~\ref{baseline} and~\ref{implementation details}.

\subsection{Deployed EGM Design Experiments} \label{deployedexperiment}

The human-AI team is tasked with designing three EGMs for deployment: an Agriculture, Nutrition, and Resilience EGM. For each one, the human-AI interaction and, therefore, our experiments are targeted at the title and abstract screening process.  

The three EGMs are created independently and following the same process. They start with the following size of datasets: Agriculture 221k, Nutrition 117k, and Resilience 60k. The human-AI workflow used for the title and abstract screening of each EGM is shown in Figure~\ref{fig:work} and described above in section~\ref{humanaiteam}. Once the process is complete for each EGM, the human-AI team has labeled a small subset of each dataset. We utilize these labeled datasets for the following experiments.

\subsubsection{Model Performance Experiments}
We explore the effectiveness of the industry standard SVM methods and our proposed BERT method in classifying documents as relevant or irrelevant for each of the three EGMs. Due to the real-world nature of this case study, we only have labels for those documents which we choose to screen. For each of these labeled datasets, we perform a 85\% - 15\% train-test-split and determine the classification accuracy. 

\subsubsection{Human Effort Experiments}
We also compared the trained BERT and SVM models in terms of human effort to assess hybrid team efficiency. Specifically, we suppose the documents in the test set would be screened in descending order of priority scores predicted by the BERT and SVM models respectively. Human effort is defined as the percentage of documents that humans need to screen for getting a specific inclusion rate. The hybrid team is more efficient if fewer documents must be screened to obtain the same number of included documents. That is, less human effort is needed.

\subsection{Simulated EGM Design Experiments} \label{simulatedexperiment}




In this case study assume the human-AI hybrid team is tasked with screening a set of 68,539 documents to identify documents satisfying a given scope. This dataset is fully labeled, and we can therefore test the efficacy of different training sizes and active learning sampling strategies.

\subsubsection{Training Size Experiments}

For ML model training, a larger training set often benefits model performance but needs more human effort to label the data. In the human-AI hybrid team, the trade-off between the model performance and the required human effort for labeling should be balanced carefully to achieve high team efficiency. We conduct experiments to investigate how the training size affects hybrid team efficiency  - both in terms of the model performance (F1 score) and human effort required. Specifically, we start with an initial training set of 1,000 papers; to expand the training set, we randomly sample 1,000  papers, label them, and add them to the training set in each iteration from 1,000 to 6,000. During training, we use 85\% of the papers in the training set to train our model and 15\% as a validation set. All the other papers compose the testing set.

\subsubsection{Active Learning Experiments}

When the AI agent samples new papers to be screened, the query strategy used affects the informativeness of the sampled papers, which further influences the ML model performance and hybrid team efficiency. We compare two different sampling strategies, LC and HP, with random sampling through experiments. For each sampling strategy, we start with the same initial training set of 1,000 papers with the random sampling case. After that, we sample 1,000 new papers using the LC or HP strategy to expand the training in each iteration. We experiment with training sizes ranging from 1,000 to 7,000 for the two sampling strategies.

\subsection{Baseline} \label{baseline}

To answer RQ1 - \emph{how much human effort can be saved when the AI agent is trained on an optimal data size?}, we compare the best case from the experiments with different training sizes with the baseline cases. In the first case, the human team works alone on the same task without any AI assistance. That is, the human agents randomly screen papers from the dataset. The second case employs a support vector machine (SVM)-based classifier, which is developed for retrieving randomized controlled trials and available in the EPPI-Reviewer software~\cite{THOMAS2021140}. For the second baseline, we also experiment with four different training sizes ranging from 1,000 to 7,000, from which the best model is used as the baseline.

To answer RQ2 - \emph{how much human effort can be further saved by enhancing the hybrid team through active learning?}, the best model from the experiments with different training sizes is used as the baseline,  where all the sampled papers are randomly selected. We compare the best models from the experiments with the LC sampling strategy and the HP sampling strategy to the baseline model, respectively. 

\subsection{Implementation Details} \label{implementation details}

In this study, our models are trained with a learning rate of $1 \times 10^{-5}$. There is a warm-up phase at the beginning of the training process, which lasts for one epoch. The experiments were performed on Intel(R) Xeon(R) W-2295 CPU @ 3.00GHz 3.00 GHz, with 18 cores and 256 GB of RAM. Model training and predicting were conducted on Nvidia RTX A5000 GPUs (single GPU per run). Each experiment is repeated five times. When the predicted uncertainties and PSs are needed to sample new papers with the LC and HP strategies, we use the mean values of the predictions from the five runs to improve the repeatability of the results.



\section{Results}

In the following sections we present the results of our experiments in both case studies: the deployed EGM design, and the simulated EGM design. We compare different ML models, training sizes, and active learning sampling strategies and report their effects on model performance and human effort. Further, we go on to discuss the limitations of our work and the future use of human-AI teams in EGM design.

\subsection{Deployed EGM Design Results}
\begin{figure}[htbp]
\centering
\includegraphics[width=0.85\columnwidth]{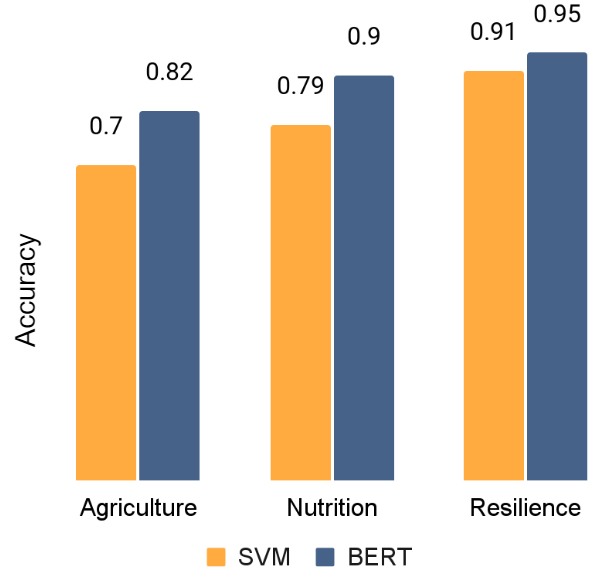}
\caption{The classification accuracy of BERT and SVM models for the three EGMs created: Agriculture, Nutrition, and Resilience.}
\label{fig:bertsvm}
\end{figure}

In this section we compare inclusion classification done by BERT and SVM models. We compare model performance across two different metrics: overall accuracy, and saved effort. The BERT model is our proposed approach, whereas the SVM model is what tools like EPPI Reviewer utilize to classify documents, and therefore represents the industry standard.

\subsubsection{Model Accuracy}

\begin{figure*}[btp]
\centering
\includegraphics[width=\textwidth]{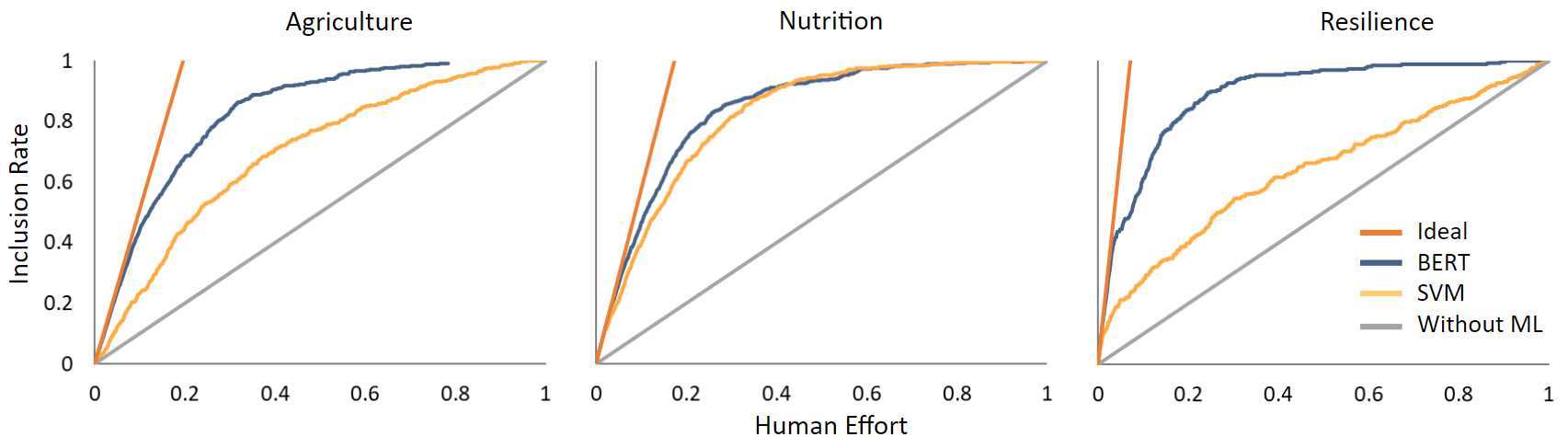}
\caption{How inclusion rate varies with human effort for the three deployed EGMs. A higher inclusion rate at a lower human effort is preferred. BERT outperforms SVM in all three EGMs.}
\label{fig:savedeffdeploy}
\end{figure*}

Figure~\ref{fig:bertsvm} shows the accuracy of the BERT and the SVM classifcation models for each of the three EGMs we created. The results show that for all three EGMs, the BERT model resulted in higher accuracy than the SVM model. The most common NLP tools used to aid EGM creation today are based on an SVM model, so our results suggest that utilizing BERT for classification has benefits over the industry standard EGM-creation tools.

\subsubsection{Saved Effort}
Figure~\ref{fig:savedeffdeploy} shows how inclusion rate varies with human effort. The orange ``Ideal'' line indicates a perfect inclusion rate, where only relevant documents are screened and therefore all documents seen are included. The grey ``Without ML'' line indicates a case in which human experts must screen all documents at random in order to find all of the included documents. We compare two ML strategies, BERT and SVM, and find that BERT outperforms SVM for all three EGMs.

We carried out a set of experiments with the screened papers of the three EGMs for the comparison. Aiming at an inclusion rate of 80\% for the screened papers, we found that the human raters needed 47\% less human effort for Agriculture, 17\% less human effort for Nutrition, and 75\% less human effort for Resilience when working with the BERT-based models rather than with the baseline EPPI-Reviewer’s SVM. The raw values of human effort for BERT and SVM for the three EGMs are shown in Table~\ref{tab:percentsaved}. The effort-saving capabilities of the BERT models are further amplified in the real screening process, in which the models are updated multiple times as new labeled documents come in as training data. In this case, the model improves iteratively over time. As its classification accuracy increases, the model can suggest only the most relevant documents to the human raters. This type of active learning is explored in the simulated dataset and described in section~\ref{al}.

\begin{table}[]
\centering
\resizebox{0.9\columnwidth}{!}{%
\begin{tabular}{c|ccc}
\textbf{EGM} & \textbf{SVM} & \textbf{BERT} & \textbf{\begin{tabular}[c]{@{}c@{}}Percent Effort \\ Saved by BERT\end{tabular}} \\ \hline
\textbf{Agriculture} & 53\% & 28\% & \textbf{47\% }\\
\textbf{Nutrition}   & 29\% & 24\% & \textbf{17\%}\\
\textbf{Resilience}  & 68\% & 17\% & \textbf{75\%}
\end{tabular}%
}
\caption{Human effort required to reach an 80\% inclusion rate for each of the three EGMs and the two ML models, SVM and BERT. The percent effort saved by BERT is calculated as the percent difference between human effort for BERT and SVM.}
\label{tab:percentsaved}
\end{table}

\subsection{Simulated EGM Design Results}

\subsubsection{Saved Effort}



\begin{figure}[ht]
\centering
\includegraphics[width=0.9\columnwidth]{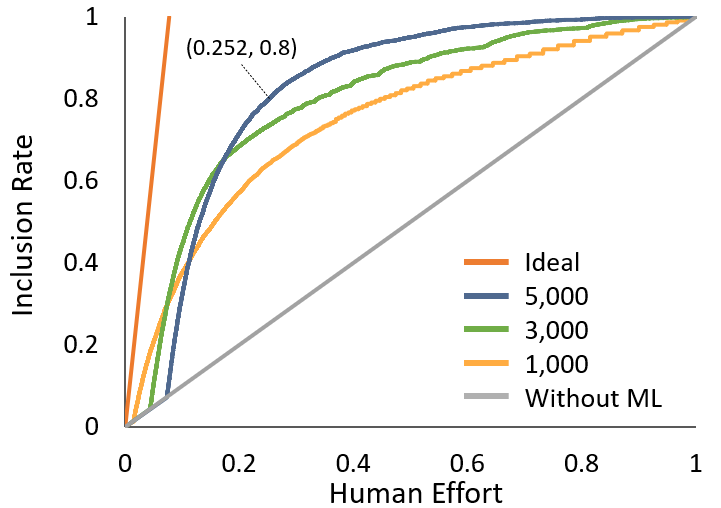}
\caption{How inclusion rate varies with human effort when our BERT-based model is trained with different training sizes. A training size of 5,000 performs best, as indicated by reaching an inclusion rate of 0.8 with the lowest human effort. }
\label{fig:randomIR}
\end{figure}
The performance of the human-AI hybrid team is assessed through inclusion rate (IR) and human effort (HE). Figure~\ref{fig:randomIR} shows the variation of IR with HE when our model is trained with different training sizes. The grey ``Without ML'' line in the figure corresponds to the condition where the human agents work alone without any AI agent. Since the human agents randomly screen papers form the dataset, IR is equal to HE in this case. The orange ``Ideal'' line close to the y-axis denotes the ideal case, in which each screened document is an included document, and no excluded documents are screened. The slope of this line is 5,281 (the total number of included papers) / 68,539 (the total number of papers in the dataset). The other curves in the figure describe how the IR changes as the human agents invest more screening efforts when the BERT-based AI agent is trained on the datasets with different sizes.

Each curve consists of two parts. The first straight line part indicates the process that the human agents label papers from the dataset to prepare the training set. Since the screened papers are randomly selected, the IR is equal to HE. Once the training set is ready, our model is trained on it to predict the PSs of the unlabeled papers. The curved part following the straight line corresponds to the process during which the human agents screen the unlabeled papers sequentially according to the predicted PSs. Since the unlabeled papers are prioritized for screening, the curves are much steeper in this second portion than in the first, which has the same slope as the ``Without ML'' line. In the curved portion, the initial slopes are close to the slope of the ideal line, then gradually decrease later on. This trend suggests that the papers with higher PSs are more likely to be identified as included papers than the papers with lower PSs, implying the effectiveness of the AI agent in prioritizing the unlabeled papers for screening.


\begin{figure}[!htb]
\centering
\includegraphics[width=\columnwidth]{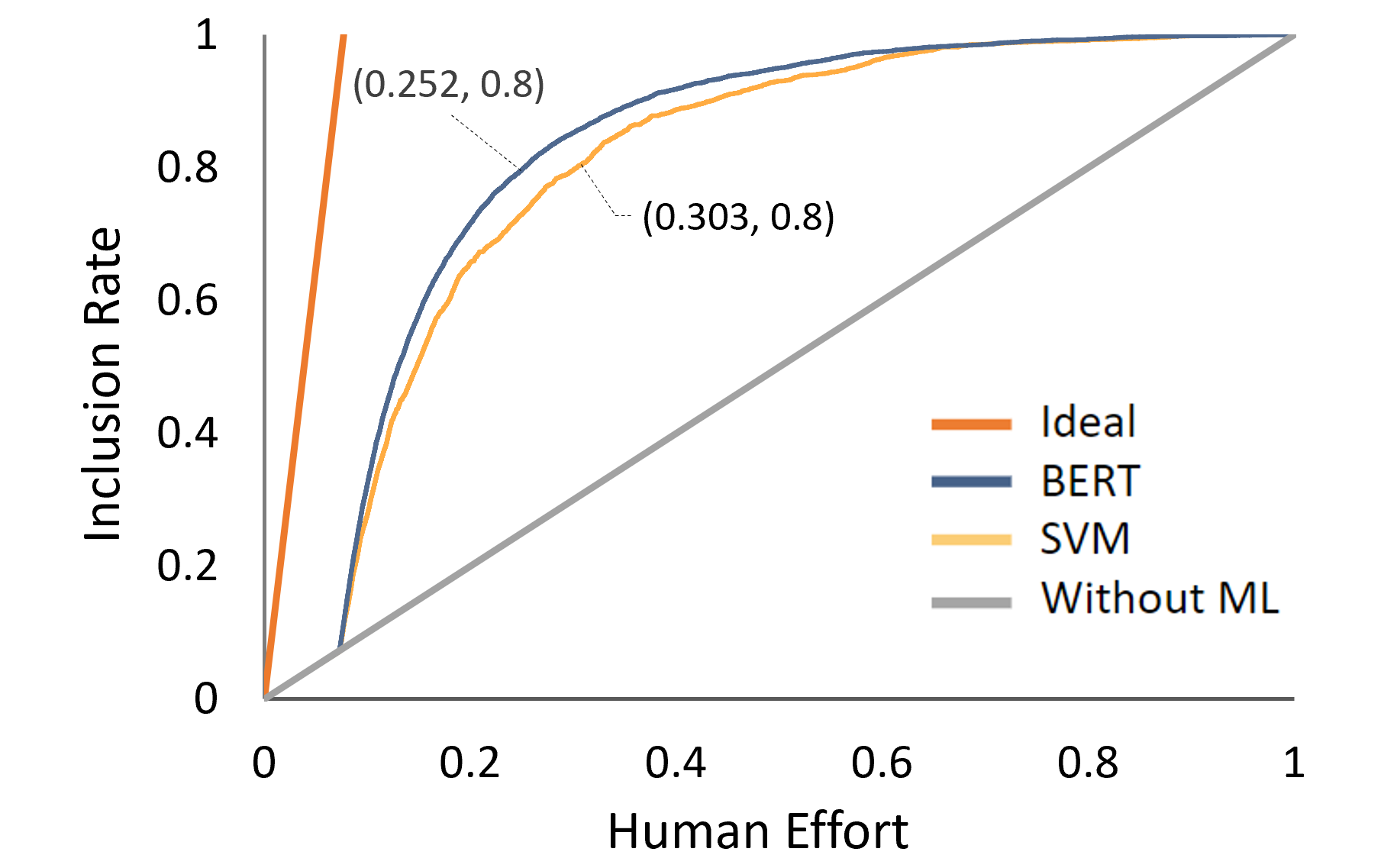}
\caption{How inclusion rate varies with human effort for the different ML models: our model (BERT), the industry standard (SVM), and ``Ideal'' and ``Without ML'' baselines.}
\label{fig:eppi_vs_bert}
\end{figure}

Since a high-performing human-AI hybrid team can achieve a higher IR with a lower HE, its initial slope should appear closer to the ``Ideal'' line in~\ref{fig:randomIR}. As the training size increases, the curve gets steeper, indicating improved model performance. This is in line with the increasing F1 scores shown in Figure~\ref{fig:F1-plot}. However, because a larger training size needs more human labeling effort (i.e., a longer straight line in the first part along the diagonal line), it may also impair the efficiency of the human-AI hybrid team. Given a target IR of 80\%, the curves show that the hybrid team gets the highest efficiency when the training size is 5,000. Under this condition, the human agents only need to screen 25.2\% of the papers to get an IR of 80\%, while they need to screen 80\% of the papers to get the same IR in the case without the AI guidance. Therefore, when the BERT-based AI agent is incorporated into the human team, it can save 54.8\% human screening effort for getting the IR of 80\%.

We also compare the BERT-based model with the SVM model used in the EPPI-Reviewer software in terms of their effectiveness as the AI agent. Similarly, we train the SVM model with different training sizes (1,000, 3,000, 5,000, 7,000), among which the training size of 5,000 needs the least human effort for getting the IR of 80\%. Figure~\ref{fig:eppi_vs_bert} compares the best BERT-based model (5,000) and the best SVM model (5,000), suggesting that the BERT-based model enables the human agents to save more screening efforts compared to the SVM model for getting any IR. Specifically, the human agents can save 5.1\% more screening efforts when working with the BERT-based AI agent than working with the SVM-based AI agent for getting the IR of 80\%. Therefore, our BERT-based model is more effective in acting as the AI agent.

\subsubsection{The Effect of Active Learning} \label{al}

In the following section, we discuss how the strategies for sampling new data to expand the training size affect the performance of the AI agent and the efficiency of the human-AI hybrid team for the TA screening task. 

\textbf{Active Learning, Training Size, and Model Performance}

\begin{figure}[!htb]
\centering
\includegraphics[width=\columnwidth]{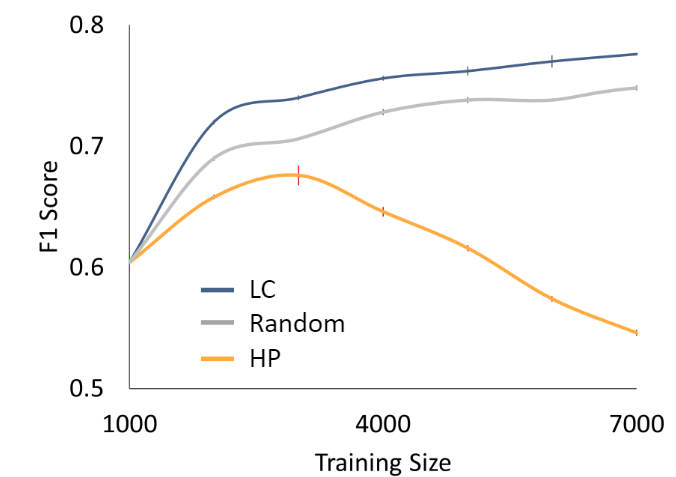}
\caption{How model performance, as shown by the F1 score, varies with training size and active learning sampling strategy. The bars indicate one standard error. We find that the least confidence (LC) strategy performs the best.}
\label{fig:F1-plot}
\end{figure}

Here we report the results of the experiments with different training sizes and different sampling strategies to demonstrate the effect of incorporating the AI agent into the human team, answering RQ1. Following the protocol of the classification problems with the imbalanced dataset, we use the F1 score computed at the default threshold of 0.5 as the classification metric. In these experiments, the sampled papers are randomly selected. The black curve in Figure~\ref{fig:F1-plot} shows the variation of the F1 score with the training size. As the training size increases, the F1 score improves with diminishing marginal effect, especially when the training size is larger than 5,000.

\begin{figure}[!htb]
\centering
\includegraphics[width=\columnwidth]{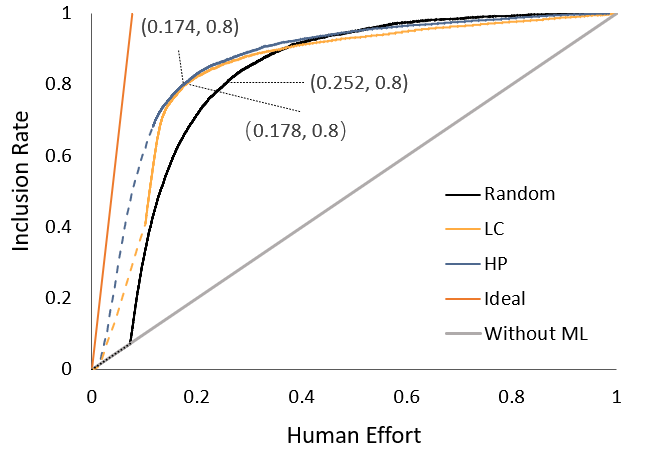}
\caption{How different AL sampling strategies affect the human effort and inclusion rate relationship. The dotted line portion of each curve represents the screening-updating-predicting-sampling iterations, while the solid line part corresponds to the process when the human agents screen the prioritized papers.}
\label{fig:sampling}
\end{figure}

\begin{figure}[!htb]
\centering
\includegraphics[width=\columnwidth]{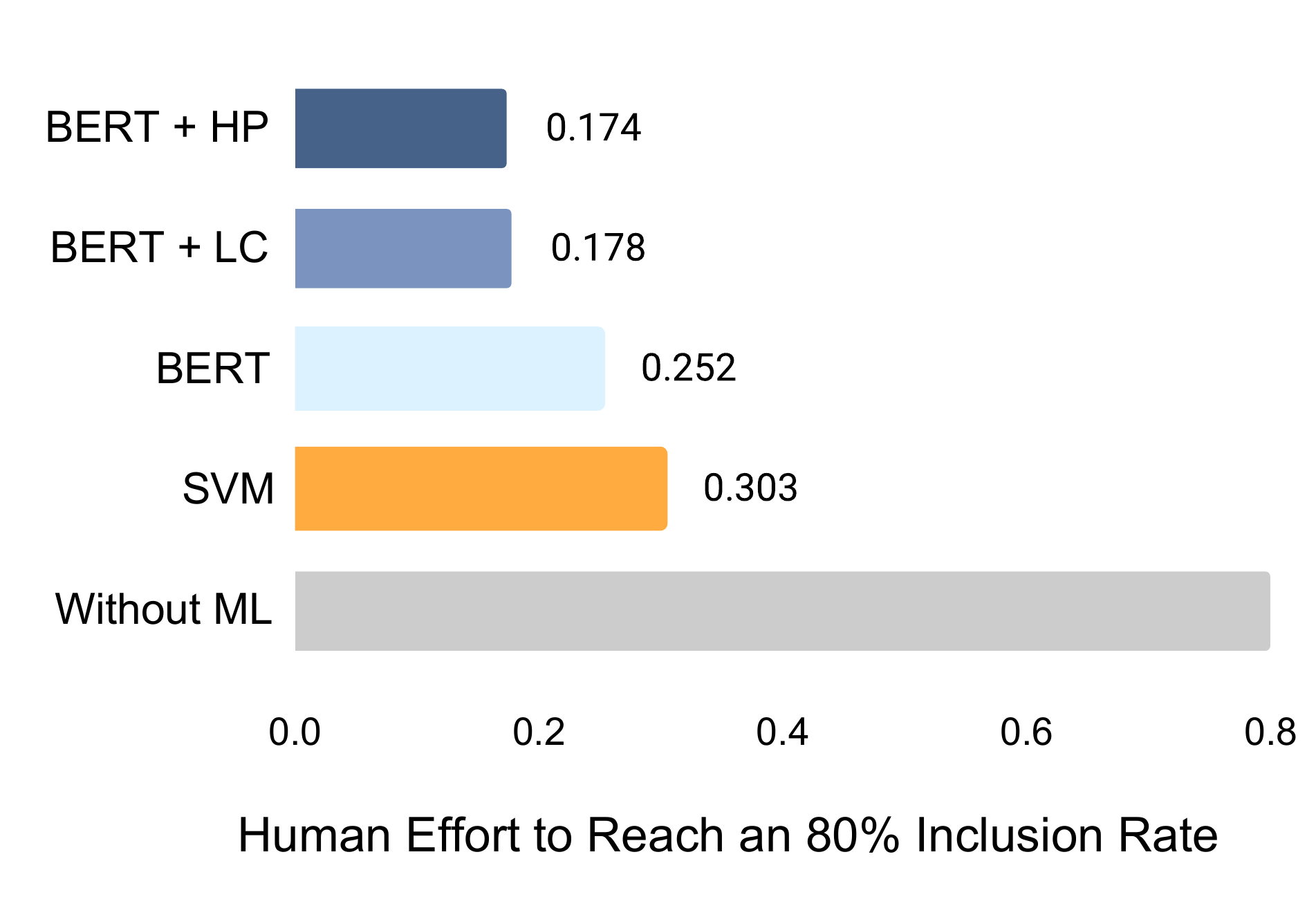}
\caption{The human effort required to reach an 80\% inclusion rate for various models. Lower human effort is preferred. With no ML, it takes 80\% human effort to reach an 80\% inclusion rate.}
\label{fig:he_bar}
\end{figure}

Similar to the random sampling case, the training size affects the performance of our classification model. As shown in Figure~\ref{fig:F1-plot}, a larger training size improves the F1 score when the LC strategy is applied. If we employ the HP strategy, a moderate training size (e.g., 2,000) benefits the F1 score most, and a larger training set impairs the F1 score when its size surpasses a certain value (e.g., 2,000). Overall, sampling new papers using the LC strategy leads to better classification models than randomly sampling new papers, as indicated by the higher F1 score; however, the HP sampling strategy results in worse classification models than random sampling, indicated by the lower F1 scores.

\textbf{Active Learning, Training Size, and Human Effort}

The selected AL sampling strategy and the training size also affect the human-AI team efficiency. Under the random sampling condition, a moderate training size can well balance the trade-off between higher model performance and more labeling effort for creating the training data, leading to the highest team efficiency. We observe similar trends for AL.
To get an IR of 80\%, the human-AI hybrid team achieves the highest team efficiency with a training size of 7,000 when the LC and HP sampling strategies are applied, respectively.

Figure~\ref{fig:sampling} compares the team efficiency among different sampling conditions, including random sampling, LC sampling, and HP sampling. We can see that the efficiency of the human-AI hybrid team is improved substantially with active learning. When the LC and HP strategies are applied, the human agents can respectively save 7.4\% and 7.8\% screening effort for getting the IR of 80\%. Specifically, the dotted line portion of each curve represents the screen-update-predict-sample iterations (i.e., the ``AI: Stop training?'' loop in Figure~\ref{fig:work}), while the solid line part corresponds to the process when the human agents screen the prioritized papers according to the predictions from the finalized AI model.

We can see that the dotted line portions of the LC and the HP curves are much steeper than the dotted line portion of the random sampling curve. The trends suggest that with both the LC and HP sampling strategies, a larger portion of the sampled papers are included papers compared to the random sampling strategy. That is, the LC and the HP sampling strategies, especially HP, improve team screening efficiency substantially during the screening-updating-predicting-sampling iterations. This can be explained by the sampling strategies themselves. The LC strategy samples the papers with the highest classification uncertainties. Given the highly imbalanced dataset, our model is less confident in classifying the papers from the minor class, i.e., the included papers from the ``1'' class, leading to more papers being sampled from the minor class. The HP strategy samples the papers with the highest predicted PSs, which are more likely to be included papers by the definition of PS.

Moreover, by sampling the papers with the highest classification uncertainties, the LC sampling strategy also enables our model to learn more efficiently from human labeling compared to the other sampling strategies. This is evidenced by the observation that the solid curve part of the blue curve is steeper at the early phase than the solid curve parts of the red and black curves in Figure~\ref{fig:sampling} and the highest F1 scores for LC in Figure~\ref{fig:F1-plot}.

\section{Discussion}

Figure~\ref{fig:he_bar} provides a comprehensive view of the various ML methods we compared, and their effect on human effort. The figure shows the human effort required to reach an 80\% inclusion rate for no ML assistance, an SVM-based model, a BERT-based model, and a BERT-based model with the LC or HP AL sampling strategies. We observe that the BERT-based model outperforms SVM, with a 16.8\% relative reduction in human effort. The results show that the AL strategies reduce human effort even further, by about 30\% compared to BERT without AL. Since our motivation is to accelerate the EGM design process and decrease the resource and time intensity of the process, this result is of great significance.

Within the hybrid team, effective interactions and mutual learning between the human agents and the AI agent can improve team performance significantly. When the LC sampling strategy is applied, both the information flow from the human agents to the AI agent (i.e., human knowledge conveyed in the labeled papers) and the information flow from the AI agent to the human agents (i.e., the AI predictions conveyed in the sampled or prioritized papers) play a role in improving the efficiency of the human-AI hybrid teams. In contrast, when the HP sampling strategy is applied, the information flow from the AI agent to the human agents plays a major role in benefiting hybrid team efficiency, especially during the screen-update-predict-sample iterations. However, in HP, the information flow from the human agents to the AI agent is not as beneficial for improving model performance.


In a practical screening process, we only know the labels of a part of the papers in a dataset, which means the actual IR, as well as its overall changing trend, is unknown. In such a scenario, it is difficult to determine when to stop expanding the training set and updating the AI agent and when to stop screening the prioritized papers. The changing scale of the predicted rankings of the unlabeled papers and the growth rate of IR can inform us about the stopping. Small changes in the paper rankings and a low growth rate of IR may suggest we stop updating the AI agent and stop screening the prioritized papers, respectively.

From the records of twelve human screeners working on an agriculture development EGM, we learn that a human screener can screen 38.6 ($SE = 1.00$ ) papers per hour on average. On this basis, the AL-enhanced AI agent can save human screeners $(80\%-17.4\%)\times68,539/38.6=1,111.5$ hours for TA screening compared to the case without the AI agent. Compared to the case using EPPT-Reviewer SVM as the AI agent, $(30.3\%-17.4\%)\times68,539/38.6=229.1$ hours can be saved by our model. 
\subsection{Discussion on AI-assisted Design of EGMs}

The nature of our work creating EGMs for deployment and use by USAID meant that our team faced many real-world challenges. In this section, we discuss the unique challenges and limitations that arise when designing EGMs with AI-assistance.

\textbf{The Cost of Communication} The cost of communicating in a human-AI team can be significant but difficult to quantify as it involves multiple factors. One such factor is the time and effort required to exchange information via email, which can lead to delays and potential miscommunication. Additionally, updating document labels and merging datasets can be a complex and time-consuming task that requires oversight and project management to ensure accuracy. These activities can also be a source of errors that can negatively impact the performance of AI models. Finally, time lags between humans labeling documents, the AI agent receiving the documents and updating the model, and then the AI agent sending back newly ranked documents for human screening means that one team may be operating with incomplete data. Therefore, optimizing communication channels and implementing efficient communication protocols can help reduce the costs associated with human-AI team collaboration.

\textbf{Trust in AI} A major challenge that many AI recommendation systems face is the ``cold start'' problem. The AI agent must provide some prediction about the documents in the first iteration, but at this point the model knows nothing about the new domain. In our case, we pre-trained our model on documents in the global development space, but this cannot ensure that it would perform well in classifying documents for, say, an Agriculture-specific EGM without any additional training data. This challenge, while common, can lead to distrust in AI from the human team, if they find the initial rankings to be incorrect. Additionally, the training data for our models are labels from people, which can be noisy. The AI model's performance is constrained by the quality of its training data, and therefore to have meaningful and accurate model results, we must begin with consistent high-quality training data.

\textbf{``When to Stop'': A Business Decision} Another challenge in the deployed EGM design case study was determining when to stop the screening process, the second question shown in Figure~\ref{fig:work}. Our human-AI team faced a trade-off here between screening more papers in order to improve model performance, or stopping screening in order to move onto the next step in the EGM process (Figure~\ref{fig:egm}). This is ultimately a business decision in which the team must weigh the resource cost of improving the model, and identifying the most ``true positive'' documents. We experimented with two techniques for determining ``when to stop.'' The first of these techniques was calculating the similarity of the rankings of the documents when ordered based on the priority score between two consecutive iterations. If the similarity of the two rankings was above a certain value after a screening iteration, we could stop updating the BERT model. The second technique was to terminate the human screening process at a specific real-time inclusion rate, e.g., the number of relevant documents identified from screening 1,000 documents in the current iteration. If the number of relevant documents is lower than a given threshold, the human screening team could stop the screening process. Future work could specifically address the question of when to stop screening, as it is a highly relevant decision for the human-AI team.

\textbf{Automation of Full-text Screening} The EGM design process, as depicted in Figure~\ref{fig:egm}, includes both title and abstract screening, and full-text screening. A natural continuation of our work would be to use NLP to assist in the full-text screening step. This step, however, presents the logistical challenge of obtaining the full-text documents. While many institutions have subscription-based access to scholarly article databases, copyright issues make downloading and using full-text documents a challenge when working among and between different institutions. This ultimately dictated that our project scope remains in the title and abstract screening process alone.

Additionally, to perform full-text screening, one would need to train another BERT model to classify full-text documents for inclusion. This means human screeners would need to generate a training dataset for this task, which would require significant human effort. Large language models (LLMs) may assist in this challenge. LLMs which are trained on billions of documents~\cite{openai2022chatgpt} have a broad understanding of language, and future work can explore whether they can classify full-text documents without domain specific training data.

\textbf{Counterfactual Analysis} Our team faced the challenge of accurately comparing different document screening techniques - such as using a BERT-based model, an SVM-based model, and no ML model within the real-world setup. It was infeasible for the human raters to create each EGM three separate times in order to compare the entire process for each technique. Therefore, we standardized our comparisons by performing retrospective experiments after the human-AI team had labeled a subset of the data. We present the results using this labeled subset. We further aimed to address this limitation by including the second case study - the simulated EGM design. In this case study, we utilized a fully labeled dataset of 68,539 documents in order to experiment with the various ML models and active learning sampling strategies. However, this challenge means we do not have true counterfactual analyses of how the EGM process would have proceeded without any AI assistance. Future work could further address this limitation by creating each EGM multiple times for each different technique.

\subsection{Future Use of Human-AI Teams for EGM Design}

The AI sub-field of NLP is experiencing rapid growth. Large language models (LLMs) like OpenAI's ChatGPT~\cite{openai2022chatgpt}, and Meta's Galactica~\cite{galactica2022} are changing the way the world perceives, exploits, and interacts with pre-trained language models. These models were released after we had concluded our EGM creation; however, we predict that their capabilities will shift the way that NLP is utilized in EGM design. 

We have performed a number of exploratory experiments to understand LLMs' capability in EGM design. We explored ChatGPT's understanding of the relationship between certain interventions and outcomes by asking it "How can agriculture transformation change poverty, migration, and food security?" The LLM captured the general qualitative relationships between the intervention (agriculture transformation) and the outcomes (poverty, migration, and food security), but did not output any quantitative implications, potential information sources, or indications of how well the relationships have been studied. This suggests to us that ChatGPT can capture the intervention-outcome relationships in a coarse resolution, but cannot provide all the detailed information that a human team or human-AI hybrid team can capture.

In a small-scale study, we explored whether LLMs could identify the interventions and outcomes of ten of the documents shown in the EGM in Figure~\ref{fig:exegm}. We fed ChatGPT the relevant abstracts and asked it for the intervention and outcome of each abstract. We found that the LLM was able to generate relevant interventions and outcomes. One such response was ``The intervention is an agriculture transformation program, which is not further specified in the paper. The outcome variables evaluated in the study are poverty, migration, food security, and agricultural revenue.'' This generative capability can be powerful in the early stage of EGM design, during which human experts determine the intervention and outcome categories that frame the EGM scope (shown as the column and row headers in Figure~\ref{fig:exegm}). Generative text can also be powerful for creating brief summaries of many documents, which falls under the overall goal of evidence synthesis.

\section{Conclusion}

In this paper, we have studied (1) how incorporating the BERT-based AI agent into the human team affects team efficiency in the EGM design process and (2) how enhancing the hybrid team through active learning can improve hybrid team efficiency. We propose a human-AI hybrid teaming workflow during TA screening portion of the EGM design process. We a) design and deploy three EGMs for global development in the areas of Agriculture, Nutrition, and Resilience, and b) conduct simulated experiments with a fully labeled dataset to answer the research questions described above. Our results show that the data size for training the AI agent influences hybrid team efficiency. When the training size is optimized, the incorporation of the BERT-based AI agent can reduce human effort by 68.5\% compared to the case without AI assistance and by 16.8\% compared to the case using an SVM-based agent for getting to an inclusion rate of 80\%. Moreover, enhancing the hybrid team through active learning can further reduce human effort by 30\% compared to BERT with no active learning. The proposed human-AI hybrid teaming workflow has been validated in the practical construction process of three EGMs. Therefore, the AL-enhanced human-AI hybrid team can accelerate evidence gap map design, and decision making in the global development field significantly.

\section*{Acknowledgments}
This publication was made possible through support provided by the USAID Bureau for Resilience and Food Security, U.S. Agency for International Development. The funders had no role in study design, data collection and analysis, decision to publish, or preparation of the manuscript. The opinions expressed herein are those of the authors and do not necessarily reflect the views of the U.S. Agency for International Development.

\bibliographystyle{asmeconf}  
\bibliography{main}

\end{document}